\documentclass[sigconf]{acmart}
\AtBeginDocument{%
  }

\copyrightyear{2025}
\acmYear{2025}
\setcopyright{cc}
\setcctype{by}
\acmConference[KDD '25]{Proceedings of the 31st ACM SIGKDD Conference on Knowledge Discovery and Data Mining V.2}{August 3--7, 2025}{Toronto, ON, Canada}
\acmBooktitle{Proceedings of the 31st ACM SIGKDD Conference on Knowledge Discovery and Data Mining V.2 (KDD '25), August 3--7, 2025, Toronto, ON, Canada}
\acmDOI{10.1145/3711896.3737234}
\acmISBN{979-8-4007-1454-2/2025/08}

\usepackage{algorithm}
\usepackage{algpseudocode}

\usepackage{graphicx}
\usepackage{subcaption}
\usepackage{multirow}
\usepackage{enumitem}
 \usepackage{balance}

\theoremstyle{acmdefinition}
\newtheorem{definition}{Definition}

\def\a{{\boldsymbol a}}

\def\b{{\boldsymbol b}}

\def\h{{\boldsymbol h}}

\def\L{{\boldsymbol L}}

\def\X{{\boldsymbol X}}

\def\EM{{\mathcal E}}

\def\GM{{\mathcal G}}

\def\VM{{\mathcal V}}

\def\TB{{\mathbb T}}

\def\1{\mathds{1}}

\begin{document}

\title{Hierarchical Structure Sharing Empowers Multi-task Heterogeneous GNNs for Customer Expansion}

\author{Xinyue Feng}
\orcid{0009-0009-5326-6818}
\affiliation{%
  \institution{Rutgers University}
  \city{Piscataway}
  \state{NJ}
  \country{USA}
}
\email{xinyue.feng@rutgers.edu}

\author{Shuxin Zhong}
\authornote{Corresponding author}
\orcid{0009-0006-1758-2870}
\affiliation{%
  \institution{HKUST(GZ)}
  \state{Guangzhou}
  \country{China}
  }
\email{shuxinzhong@hkust-gz.edu.cn}

\author{Jinquan Hang}
\orcid{0000-0002-2547-5614}
\affiliation{%
  \institution{Rutgers University}
  \city{Piscataway}
  \state{NJ}
  \country{USA}
  }
\email{jinquan.hang@rutgers.edu}

\author{Wenjun Lyu}
\orcid{0000-0002-7885-3105}
\affiliation{%
  \institution{Rutgers University}
  \city{Piscataway}
  \state{NJ}
  \country{USA}
  }
\email{wenjun.lyu@rutgers.edu}

\author{Yuequn Zhang}
\orcid{0009-0006-7906-9132}
\affiliation{%
  \institution{Rutgers University}
  \city{Piscataway}
  \state{NJ}
  \country{USA}
  }
\email{yz1127@cs.rutgers.edu}

\author{Guang Yang}
\orcid{0009-0001-2364-0188}
\affiliation{%
  \institution{Rutgers University}
  \city{Piscataway}
  \state{NJ}
  \country{USA}
  }
\email{gy121@cs.rutgers.edu}

\author{Haotian Wang}
\orcid{0000-0001-9783-6389}
\affiliation{%
  \institution{JD Logistics}
  \state{Beijing}
  \country{China}
  }
\email{wanghaotian18@jd.com}

\author{Desheng Zhang}
\orcid{0000-0001-9307-8736}
\affiliation{%
 \institution{Rutgers University}
 \city{Piscataway}
  \state{NJ}
  \country{USA}
 }
 \email{desheng@cs.rutgers.edu}

\author{Guang Wang}
\orcid{0000-0002-7739-7945}
\affiliation{%
 \institution{Florida State University}
 \city{Tallahassee}
  \state{FL}
  \country{USA}
 }
\email{guang@cs.fsu.edu}


\renewcommand{\shortauthors}{Xinyue Feng et al.}


\begin{abstract}
Customer expansion, i.e., growing a business’s existing customer base by acquiring new customers, is critical for scaling operations and sustaining the long-term profitability of logistics companies. 
Although state-of-the-art works model this task as a single-node classification problem under a heterogeneous graph learning framework and achieve good performance, they struggle with extremely positive label sparsity issues in our scenario. Multi-task learning (MTL) offers a promising solution by introducing a correlated, label-rich task to enhance the label-sparse task prediction through knowledge sharing.
However, existing MTL methods result in performance degradation because they fail to discriminate task-shared and task-specific structural patterns across tasks. This issue arises from their limited consideration of the inherently complex structure learning process of heterogeneous graph neural networks, which involves the multi-layer aggregation of multi-type relations.
To address the challenge, we propose a Structure-Aware Hierarchical Information Sharing Framework (SrucHIS), which explicitly regulates structural information sharing across tasks in logistics customer expansion. SrucHIS breaks down the structure learning phase into multiple stages and introduces sharing mechanisms at each stage, effectively mitigating the influence of task-specific structural patterns during each stage.
We evaluate StrucHIS on both private and public datasets, achieving a 51.41\% average precision improvement on the private dataset and a 10.52\% macro F1 gain on the public dataset. StrucHIS is further deployed at one of the largest logistics companies in China and demonstrates a 41.67\% improvement in the success contract-signing rate over existing strategies, generating over 453K new orders within just two months.

\end{abstract}



\maketitle

\newcommand\kddavailabilityurl{https://doi.org/10.5281/zenodo.15567831}

\ifdefempty{\kddavailabilityurl}{}{
\begingroup\small\noindent\raggedright\textbf{KDD Availability Link:}\\
The source code of this paper has been made publicly available at \url{\kddavailabilityurl}.
\endgroup
}

\section{Introduction} \label{sec:intro}

\begin{figure}[!t]
\centering
\includegraphics[scale=.8]{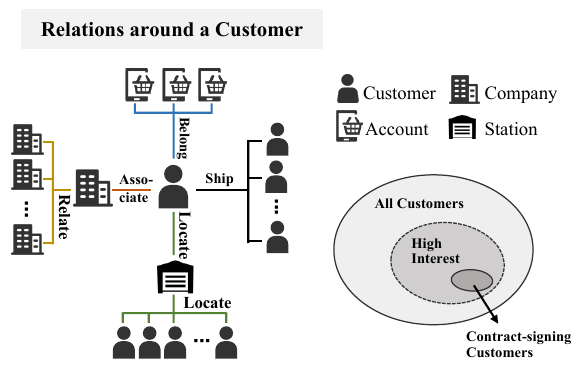}
\caption{The figure shows various relations around the customer in our logistics scenario. Our goal is to identify potential customers willing to establish contractual partnerships for long-term logistics services.} 
\label{fig:bg}
\end{figure}

The proliferation of online shopping has exponentially increased logistics demands, driving growth for logistics companies like Amazon~\cite{Amazon}, Fedex~\cite{Fedex}, JD Logistics~\cite{JD}, and SF Express~\cite{SF}. 
This competitive landscape compels logistics companies to increasingly focus on \textbf{customer expansion}~\cite{Hubble, ad2, ad3, AITM} - seeking high-value customers willing to sign long-term contracts. 
Establishing such partnerships not only ensures stable revenue streams but also reduces customer churn risks posed by business competitors, making it become a key strategic priority for the industry.

Due to its importance, companies and researchers have explored various approaches to effectively identify potential customers for expansion \cite{ad1,ad2,ad3}. 
The most recent methods~\cite{Hubble, AD-temporalgraph} formulate customer expansion as a single node classification problem in heterogeneous graph learning~\cite{HGN-Survey}, allowing them to capture complex structural patterns of customers (i.e. relations between customers and other entities) to improve predictive capabilities.
However, they struggle with our scenario, where the positive label (i.e., the number of contracted customers) is extremely sparse.
The severe positive label sparsity limits models’ ability to extract meaningful structural patterns to discriminate potential customers, consequently compromising predictive performance~\cite{labelsparsity}.

A promising approach to address this problem is using multi-task learning (MTL) \cite{caruana1997multitask} by introducing high-interest customer prediction task, which provides abundant positive labels and exhibits a natural correlation with the contract-signing customer prediction task, as all contract-signing customers originate from high-interest customers. 
The inherent relationship between these tasks makes MTL particularly effective, as the knowledge gained from the label-rich task can enhance the performance of the label-sparse prediction task~\cite{AITM,guobs,LimHNGWT22}, thereby alleviating the positive data scarcity issue.

However, our experiment results reveal that directly applying MTL for cross-task knowledge sharing deteriorates performance (see Section \ref{sec:limitation} for details). 
The key issue lies in that some structural patterns benefit all tasks, while others are task-specific (Figure~\ref{fig:challenge}), yet the shared backbone design in MTL \cite{Boosting, LZZLD021,SaravanouTML21,LimHNGWT22} indiscriminately shares all structural patterns.
Advanced MTL solutions on non-graph data~\cite{MMoE,PLE,MultiSFS}, motivated by Mixture of Experts (MoE)\cite{moe-survey} framework, employ multiple backbones to implicitly learn task-specific or task-shared patterns separately. 
While effective for non-graph scenarios, these \textbf{implicit} discrimination methods struggle with structural patterns in heterogeneous graph (Figure~\ref{fig:challeng_moe}), which involves multi-layer aggregation of multi-type relations.

To this end, we propose a \textbf{Structure-Aware Hierarchical Information Sharing Framework (SrucHIS)}, which \textbf{explicitly} regulates structural information sharing across different tasks in logistics customer expansion, e.g., high-interest customer prediction and contract-signing customer prediction tasks. 
Specifically, rather than following a sequential process where heterogeneous graph neural networks (HGNNs) first learn all structural information before enabling cross-task sharing, SrucHIS hierarchically decomposes the structure learning process into distinct stages and integrates information sharing mechanisms at each stage, enabling concurrent structural learning and inter-task information interaction. 
Our sharing mechanism is implemented in two levels: i) relation-wise sharing, which selectively exchanges structural information for specific relations, and ii) layer-wise sharing, which enables selective sharing of multi-hop structural information.
By such design, SrucHIS ensures that task-specific requirements are addressed during each step of structural learning, easily and effectively mitigating the influence of task-specific structural patterns.
Our main contributions are summarized as follows:

\begin{itemize}[leftmargin=*]
    \item We observe and analyze a critical yet previously under-researched issue based on our large-scale industry data: implicit cross-task structural interference in customer expansion. Through experiments, we highlight that existing MTL approaches suffer from unintended structural interference between tasks, leading to degraded performance.  
    \item 
    We introduce SrucHIS, a heterogeneous graph-based MTL framework that explicitly regulates structural knowledge sharing to mitigate cross-task interference. 
    SrucHIS breaks down the structure learning phase into multiple stages and introduces sharing mechanisms at each stage, ensuring that task-specific requirements are addressed during each stage.
    \item
    We implement, evaluate, and deploy SrucHIS in JD Logistics, one of the largest logistics companies in China.  
    To assess the effectiveness of our method, we conduct extensive offline experiments using three months of real-world data. Compared with state-of-the-art MTL methods, our approach effectively mitigates the performance drop caused by cross-task interference, achieving a 51.41\% improvement in Average Precision.  
    To further evaluate the robustness and generalization of our method, we test it on two public datasets, where it achieves a 10.52\% improvement in macro F1 scores.  
    Finally, in real-world deployment, StrucHIS demonstrate a 41.67\% improvement over existing strategies, generating over 453K new orders within just two months, highlighting its substantial commercial value. 
\end{itemize}

\section{Preliminaries}

\subsection{Definitions}
\begin{definition}
\textbf{Logistics Customer Expansion.}
Our logistics customer expansion process inherently contains two stages: 
(1)~\textit{Telephone screenings} identify~\textit{high-interest customers}, who make up 2\% of the total customer base and are potentially interested in our logistics services. 
(2)~\textit{In-person visits} negotiate shipping prices, package volume, and finalize contracts with~\textit{contract-signing customers}, who only make up 0.07\% of the total customer base
Notably, all contract-signing customers originate from high-interest customers, which is 20 times more numerous. 

\end{definition}

\begin{definition}
\textbf{Multi-task Learning.}
Multi-task learning (MTL) is a learning paradigm where multiple tasks are trained simultaneously, leveraging shared information across these tasks to improve the overall performance compared to training each task independently~\cite{caruana1997multitask, survey-mtl}. 
\end{definition}

\begin{definition}
\textbf{Heterogeneous Graph.}
A heterogeneous graph is defined as $\GM=\{\VM, \EM, \tau, \psi\}$, where $\VM$ and $\EM$ denote the sets of nodes and edges, respectively.
Each node $v\in \VM$ has a type $\tau(v)$, forming the node type set $\TB_{\VM} = \{\tau(v) : \forall v \in \VM\}$. 
Each edge $e\in \EM$ has a type $\psi(e)$, forming the edge type set $ \TB_{\EM} = \{\psi(e) : \forall e \in \EM\}$.
For a heterogeneous graph, $|\TB_{\VM}| + |\TB_{\EM}| \geq 2$. If $|\TB_{\VM}| = |\TB_{\EM}| = 1$, it will degenerate to a homogeneous graph~\cite{HGB}.  
\end{definition}

\subsection{Problem Formulation}
We formalize the logistics customer expansion problem as a classification task within a heterogeneous graph-based multi-task framework. 
Entities such as customers and companies, along with their connections, are modeled as a heterogeneous graph $\GM$ as we defined in Def.1. \textbf{In this paper, the terms "relation" and "edge" are used interchangeably to refer to the connections within the graph.}
Node and edge feature sets are denoted by  $\X_{\VM} = \{x_v|v \in \VM \}$ and $\X_{\EM} = \{x_e|e \in \EM \}$, respectively.
We aim to predict two binary-classification tasks: (1) \textit{\textbf{Task 1: High-interest Customer Prediction}} (interested or not interested), (2) \textit{\textbf{Task 2: Contract-Signing Customer Prediction}} (successfully sign-up conrtact or unsuccessfully sign-up contract). 
Let the target customer node set be $\{\tilde{v}|\tilde{v}\in \tilde{\VM}\}$, where $y_{\tilde{v}}^t$ denotes the label of $t^{th}$ task for node $\tilde{v}$.
Our goal is to train a MTL framework that jointly optimizes both two tasks  $(y_{\tilde{v}}^1, y_{\tilde{v}}^2)=F(\GM; \X_{\VM}; \X_{\EM}; \theta)$.

\begin{figure}[!t]
\centering
\includegraphics[scale=.78]{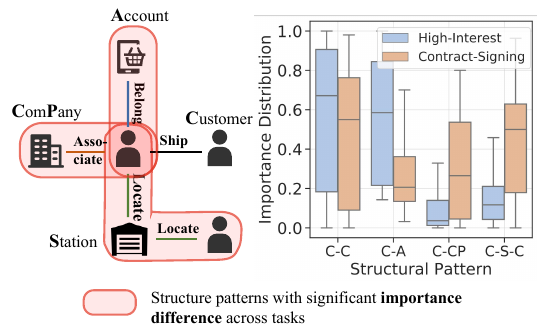}
\caption{The importance distribution of structural patterns learned by different tasks, where '\textit{C-C}' denotes the 1-hop structural pattern '\textit{Customer-(ship)-Customer}', '\textit{C-S-C}' denotes the 2-hop structural pattern '\textit{Customer-(locate)-Station-(locate)-Customer}', etc. }
\label{fig:challenge}
\end{figure}

\subsection{Performance Degradation in Our Scenario}\label{sec:limitation}

In this section, we validate our hypothesis that \textit{some structural patterns benefit all tasks, while others are task-specific, leading to performance degradation in MTL.}  
Specifically, we:  
(1) identify and quantify the varying impact of a specific structural pattern on different tasks; and  
(2) show that existing implicit disentanglement mechanisms fail to effectively discriminate task-shared and task-specific structures.

\textbf{The varying impact of a specific structural pattern on different tasks.} \; 
We quantify the importance of a structural pattern for a task by the attention weights between nodes connected by the structural pattern. 
To make the visualization result more convincing, we collect weights from all relevant node pairs, and form them as an importance distribution of this type of structural pattern. 
For example, the importance of 1-hop structural pattern '\textit{Customer-(ship)-Customer}' is the attention weight when aggregating a customer node to another customer node connected by 'ship' relation in the first heterogeneous graph neural network later. 
Similarly, the importance of 2-hop relation '\textit{Customer-(locate)-Station-(locate)-Customer}' is obtained by the attention weights when aggregating a customer node to a station node connected by 'locate' relation from the second heterogeneous graph neural network layer. 
As shown in Figure \ref{fig:challenge}, though structural pattern '\textit{Customer-(ship)-Customer}' is important for both tasks, structural pattern  '\textit{Account-(belong)-Customer}' is significantly more important for Task 1 than for Task 2, which verifies our claim.

\begin{figure}[!t] 
    \centering
    \captionsetup[subfigure]{aboveskip=-0.5pt}
    \begin{subfigure}[b]{0.21\textwidth}
        \centering
        \includegraphics[width=\textwidth]{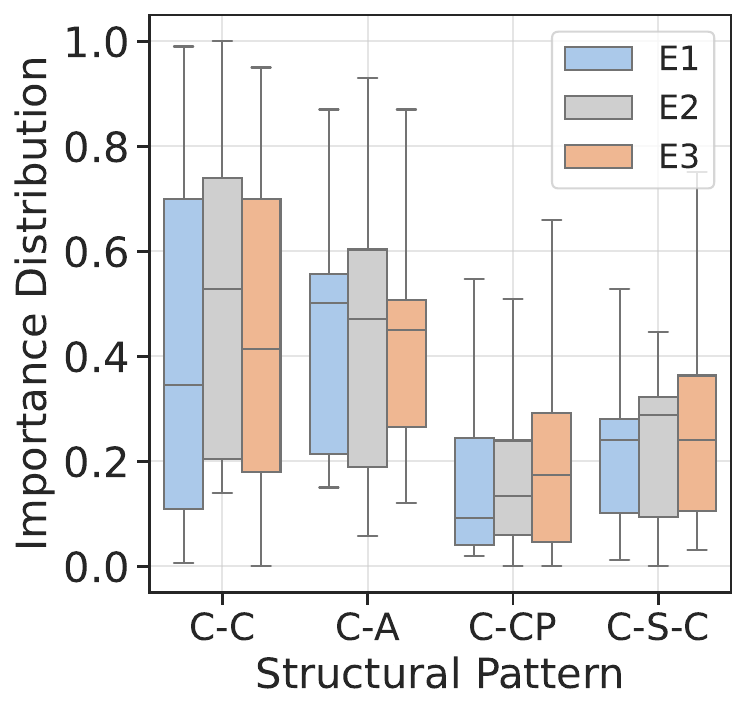}
        \caption{MoE MTL Method}
    \end{subfigure}
    \hspace{0.9mm} 
    \begin{subfigure}[b]{0.21\textwidth}
        \centering
        \includegraphics[width=\textwidth]{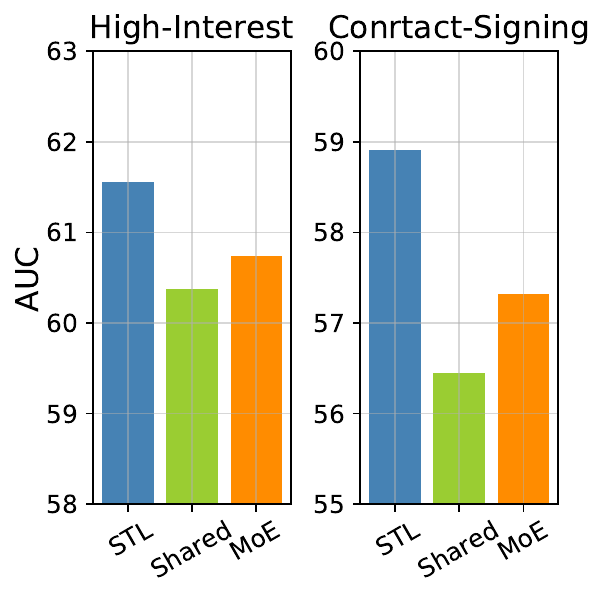}
        \caption{Performance Comparison}
    \end{subfigure}
\caption{(a) The importance distribution of structural patterns learned by different experts (E1, E2, and E3) in MoE-based MTL method PLE.  (b) Performance comparison of single-task learning, Shared-backbone MTL method, and MoE-based MTL method.}
\label{fig:challeng_moe}
\end{figure}

\textbf{Challenges in existing MTL frameworks.}\;
Most MTL methods use a shared backbone for all tasks to ensure information sharing between tasks, with separate prediction heads for each task ~\cite{AITM, guobs, WangLLYW20,LZZLD021,SaravanouTML21,LimHNGWT22}. 
In our scenarios, a single heterogeneous graph neural network (HGNN) is used for both tasks, making all tasks learn identical attention weights between two nodes. However, our visualization in \ref{fig:challenge} has revealed that different tasks may exhibit distinct preferences for the same type of structural pattern. Thus, learning identical weight for such a structural pattern for all tasks inevitably hampers task-specific performance (Figure \ref{fig:challeng_moe}(b)).

Motivated by advanced MTL methods on non-graph data~\cite{MMoE,PLE,MultiSFS}, a solution is to introduce multiple backbones (e.g., Mixture of Experts (MoE)\cite{moe-survey}) into MTL to implicitly learn task-specific and task-shared structural patterns separately.
Specifically, we can use multiple experts (HGNNs) to extract diverse information and gate networks for selective information exchange.
Despite these methods implicitly learn to distinguish shared and specific information in non-graph scenarios, they neglect that structural information is deeply entangled in embeddings through multi-step aggregation across multi-type relations.
For example, in a 1-hop aggregation setting, although contract-signing prediction benefits from the \textit{C-C} structure, its contribution remains entangled with other relations.
This issue becomes even more severe in multi-hop aggregation, where information from multiple structural patterns is compressed into a single embedding.
As a result, when applied to heterogeneous graphs, these methods struggle to effectively distinguish between task-shared and task-specific structural information.  

We support the claim by Figure \ref{fig:challeng_moe}, where we show the structural pattern importance learned by different experts (E1, E2, and E3) in the PLE model~\cite{PLE}. Ideally, E1 and E3 should focus on learning task-specific structural patterns, while E2 should capture commonalities across tasks. However, the figure shows that all experts tend to learn similar and ‘average’ importance for each structural pattern, which demonstrates that PLE struggles to selectively extract task-specific structural information.

\textbf{Motivation.}\; 
Thus, rather than implicitly learning to selectively share information, our observation motivates us to explicitly regulate structural information sharing across tasks.
Specifically, SrucHIS breaks down the structure learning phase in HGNN into multiple stages and introduces sharing mechanisms at each stage. Given an input node, (1) The HGNN first aggregates information from neighboring nodes based on relation types. 
Visualizations reveal that the importance of a relation, such as the 1-hop relation \textit{Account-(belong)-Customer}, may vary significantly across tasks.
To ensure that task-specific structures are preserved while enabling effective knowledge transfer, we selectively share only those relations that are beneficial for both tasks within each layer, rather than indiscriminately sharing all relational structures around a node.
(2) Then, the HGNN then repeats this process to capture multi-hop structural information. As visualizations show that structural preferences also vary across tasks at deeper hops, we extend the above information sharing mechanism across multiple layers of the graph, ensuring that multi-hop structural knowledge is integrated while minimizing task interference at each hop.

\section{METHODOLOGY}

\subsection{Model Overview}
\begin{figure}[t]
\centering
\includegraphics[scale=0.7]{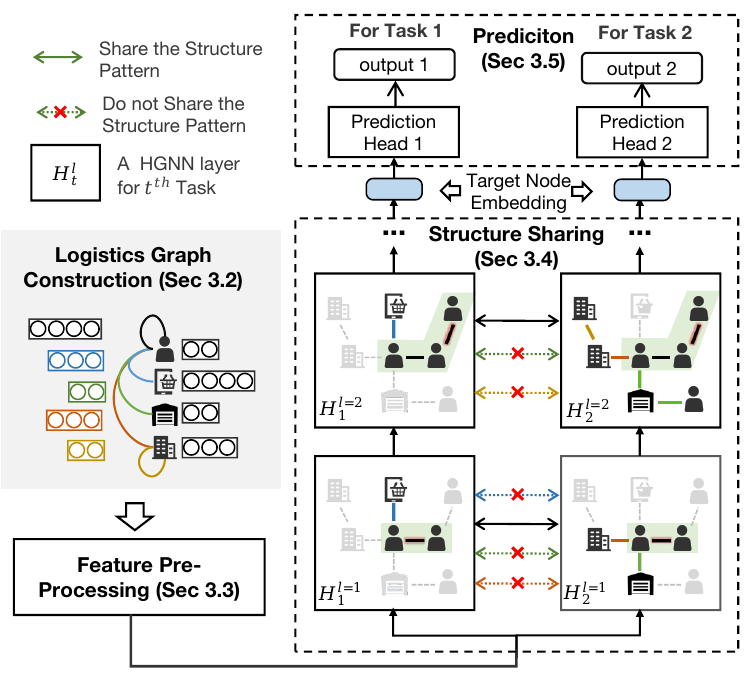}
\caption{Overview of the framework. The light green region highlights shared structural patterns between two tasks.}
\label{fig:model}
\end{figure}

Figure \ref{fig:model} shows the overview of our StrucHIS framework, which mainly consists of the following four parts:
\begin{itemize}[leftmargin=*]
    \item \textbf{Logistics Heterogeneous Graph Construction:} We construct the graph by modeling entities (customer, company, account, and station) as nodes and their interactions as edges. Each node and edge is enriched with unique feature representations.
    \item \textbf{Feature Pre-Processing:} To ensure compatibility across diverse node and edge types, we project all feature types into the identical embedding length.
    \item \textbf{Structural Information Sharing:} As the core of our framework, this module decomposes the structural learning process of HGNNs into stages and enables selective information sharing at each stage. The pseudo code can be found in Appendix \ref{sec:app_method}.
    \item \textbf{Prediction:} After L layers of structural learning and selective sharing, the refined embedding of the target customer node is fed into the prediction head to facilitate task-specific prediction.
\end{itemize}

\subsection{Logistics Graph Construction}\label{sec:data}
\textbf{How to construct the logistics heterogeneous graph?}\;
As shown in Figure \ref{fig:bg}, the logistics heterogeneous graph consists of four types of nodes, i.e., Customer, Company, Station, and Account.  
These nodes are connected by five types of edges (relations): 
(1) \textit{Associate}: a customer is the legal representative of a company, 
(2) \textit{Ship}: a customer used to send or receive packages from a customer, (3) \textit{Belong}: an online shopping account is belong to a customer. 
(4) \textit{Relate}: a company is a child/parent company of a company.
(5) \textit{Locate}: a customer is located near a delivery station. 
Note that to enhance privacy, all key identifiers (e.g., customers and accounts) have been encrypted, we only keep the encrypted ID in the graph.
All nodes and edges have their own attributes, for instance, a customer's feature attributes its total number of shipments through different channels in the past 1/3/6/12 months. Company nodes include attributes reflecting the company’s size and its industry sector, and the “Ship” relation includes attributes about the volume of shipments exchanged between the two companies.

\textbf{Why we construct the heterogeneous graph and utilize heterogeneous graph learning?}\;
A customer's likelihood of contracting is influenced by both individual attributes and its broader relational context.
For instance, the contract-signing rate at a local express station reflects regional demand, while package exchanges between companies indicate business activity and engagement.
Modeling these interactions with a heterogeneous graph enables the automatic extraction of complex, long-range dependencies, enhancing predictive accuracy.

\begin{figure*}[t]
\centering
\includegraphics[scale=0.75]{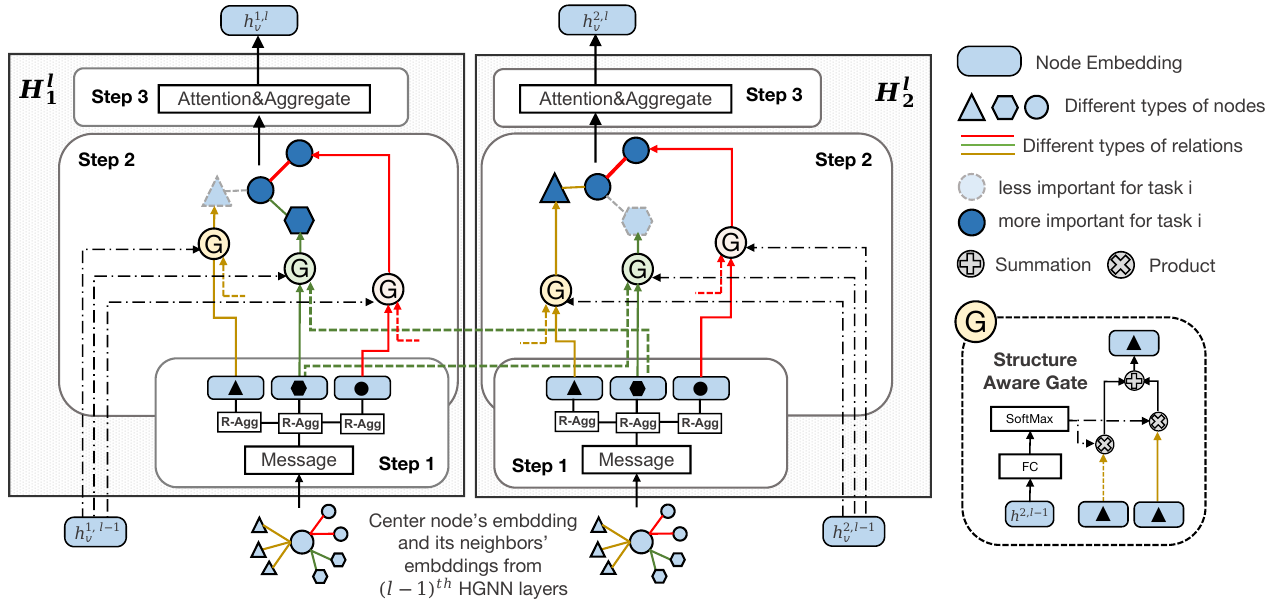}
\caption{Details of our structural information sharing module. }
\label{fig:model_detail}
\end{figure*}

\subsection{Feature Pre-processing}
Given an input heterogeneous graph, as different types of nodes and edges have different feature lengths, we use linear layers to map them into the same length. 
Let $x_v$ and $x_e$ denote node and edge feature vectors, respectively. 
$W_{\tau(v)}$ and $b_{\tau(v)}$ are the weight matrix and the bias for specific node type $\tau(v)$, while $W_{\psi(e)}$ and $b_{\psi(e)}$ are for edge type $\psi(e)$. 
Then, node and edge features are embedded by
$h_v = W_{\tau(v)}x_v + b_{\tau(v)}$ and $h_e = W_{\psi(e)}x_e + b_{\psi(e)}$.

\subsection{Structural Information Sharing Module}
\textbf{Motivation:}\;
Consider a single-layer HGNN where nodes aggregate information only from their 1-hop neighbors. Figure \ref{fig:challenge} shows that 1-hop structural pattern may hold different importance for different tasks. 
For example, the \textit{Customer-(ship)-Customer} pattern plays an important role in both tasks, whereas the \textit{Account-(belong)-Customer} pattern is significantly more relevant to high-interest prediction than to contract-signing prediction.  
This suggests that when aggregating information for a central customer node, information sharing should be encouraged between neighboring customer nodes across tasks while minimizing interactions between neighboring account nodes.

However, manually selecting which structural patterns to share is impractical, as their relevance varies across datasets. To address this, we propose an automatic structural information-sharing mechanism (Figure \ref{fig:model_detail}):

\textbf{Step 1: Relation-wise Aggregation.}\;
This step extracts and aggregates structural features from neighboring nodes based on their specific relations with the central node.
Let $h_v$ and $\{h_u, e_{v,u}\in E\}$ denote embeddings of central node $v$ and its neighboring nodes, respectively.
In the first HGNN layer, we extract task-relevant features from neighboring nodes through linear transformations, a process denoted as  \textit{Message Passing}~\cite{HGT}:
\begin{equation}
\begin{aligned}
\hat{h_u}=\mathop{Mes}(h_u, \psi(e)) = W_{\psi(e),\tau(u)}^{Mes}\cdot h_u
\end{aligned}
\end{equation}

Once the neighboring node features are transformed, we perform relation-specific aggregation. For each edge type $\psi_i$, we compute a weighted sum of the embeddings of all neighboring nodes connected to the central node via that edge type:
\begin{equation}
\begin{aligned}
h_{\psi_i} = \sum\limits_{\forall u, \psi(e_{v,u})=\psi_i}\alpha_u \hat{h_u}
\end{aligned}
\end{equation}
\begin{equation}
\begin{aligned}
[...,\alpha_u,...] = \mathop{Softmax}\limits_{\forall u, \psi(e_{v,u})=\psi_i} \left( \a_{\psi_i}^T \cdot  W^{\text{R-Agg}}_{\psi_i} [h_v \|h_u \| h_{e_{v,u}} ] \right) 
\end{aligned}
\end{equation}

\textbf{Step 2: Selective Information Sharing.}\;
After extracting structural features for each relation type, the next step is to selectively share these features across tasks. 
Given total $T$ tasks, $h_{\psi_i}^{t}$ denotes the embedding containing structural information of relation type $\psi_i$ for task $t$. For a relation type $\psi_i$, our goal is to decide whether structural information from other tasks should be shared with a target task.

To achieve this, we introduce the \textbf{Structure-Aware Gate}, motivated by the gate mechanism used in MoE~\cite{moe-survey}, which automatically decides the contribution of knowledge from different experts for a target sample.
The Structure-Aware Gate is both task-specific and relation-specific.
For a target task $t$, the gate network $G_{psi_i}^t$ computes a weighted sum of embeddings from all tasks, allowing task $t$ to leverage beneficial structural information from other tasks while suppressing interference. 
The structure-aware gate contains two steps: \textbf{weight generation} and \textbf{combination}. Here \textbf{we use task 1 as the target task} to illustrate its functionality:

1) \textbf{Weight Generation.} 
In this step, the gate network aims to generate $T$ weights for $\{h_{\psi_i}^1, ..., h_{\psi_i}^T\}$.
We use the central node embedding $\h_v^1$ as the selector and input it into a fully-connected (FC) layer followed by a Softmax function to generate weights. 
Our weight generation step is formulated as:
\begin{equation} \label{eq:our_gate_weight}
\begin{aligned}
[w^1,...,w^T] \leftarrow \mathop{Softmax}(\text{FC}^1_{\psi_i}(h_v^{1}))
\end{aligned}
\end{equation}
2) \textbf{Combination}. 
In this step, the gate network fuses $T$ structural embeddings from other tasks to the target task. 
\begin{equation} \label{eq:our_gate_single}
\begin{aligned}
  \overline{h_{\psi_i}^{1}} = \sum_{t=1..T} w^t h_{\psi_i}^{t}
  \end{aligned}
\end{equation}

In our gating mechanism, the weight generation process utilizes only the representation of the central node from each individual task, without explicitly incorporating representations from other tasks. This choice is justified because both representations of the central node from task1 and task2 are deterministic transformations of the same initial input feature , the relationship between them is consistent and learnable. During training, the gating network receives gradients influenced by both task 1 and task 2 outputs, allowing it to learn to capture this relationship. Therefore, at inference time, even only task1’s representation is used as input, the gate can still effectively predict how to use task2's information.

\textbf{Step 3: Aggregation Across Relations.}\;
For the target task, the number of structure-aware gates is the number of relation types $|\TB_{\EM}|$, where each gate generates an embedding that captures relation-specific structural information enhanced through information exchange.
We then aggregate these embeddings together to the central node and get the output embedding $h^1_v$ of the layer:
\begin{equation}\label{eq:our_final}
    h^1_v = \sum\limits_{\forall v\in V, \forall \psi_i \in \TB_{\EM}}\left(\beta_{\psi_i} \cdot \overline{h_{\psi_i}^{1}} \right)
\end{equation}
\begin{equation}
[..., \beta_{\psi_i},...] = 
\mathop{Softmax}\limits_{\psi_i\in \TB_{\EM}} \left( \b^T \cdot  W^{\text{Agg}}[h_v^1 \|h_{\psi_i}^1 ] \right) 
\end{equation}

\textbf{Apply Step 1/2/3 at Each Layer:}\;
If we only apply the above information sharing module at the first HGNN layer, it only selectively share structural information within 1-hop neighboring nodes, neglecting structural preference difference across tasks appearing at deeper hops (as shown in Figure \ref{fig:challenge}).
Thus, we apply Step 1/2/3 at each HGNN layer, ensuring that multi-hop structural knowledge is integrated while minimizing task interference at each hop.

\subsection{Multi-task Prediction}
After applying $L$ information sharing layers, to obtain the prediction result of task $t$ on target node $\tilde{v}$, we get target nodes' embeddings $h_{\tilde{v}}^{t,L}$ from the final layer and input them into task-specific prediction head $f_t$, which contains multiple fully-connected layers. Then, the objective function is constructed as 
\begin{equation}\label{eq:pred}
 L(\theta) = \sum_{t=1}^T\sum_{\tilde{v}} \L_t(y^t_{\tilde{v}}, \widehat{y^t_{\tilde{v}}}),\;
    \widehat{y^t_{\tilde{v}}} = f_t(h^{t,L}_{\tilde{v}}).
\end{equation}
where the task-specific loss $L_t$ is defined based on the task itself. 
We use the cross entropy loss for the single-label classification task and the binary cross entropy loss for multi-label classification.

\section{Experiments}

\subsection{Datasets} \label{sec:dataset}
We first utilize a large-scale offline dataset from JD Logistics to evaluate our model, and then use two other widely used public datasets in the heterogeneous graph learning field \cite{HGB, HetSANN, DiffMG} to verify the generalizability of our method. Their statistics are listed in Table \ref{tb:dataset_info}. More details are shown in Appendix \ref{app_dataset}.

\textbf{Logistics Dataset.} It is a large-scale private dataset collected from JD logistics.
For offline experiments, we collected 670,293 labeled customers based on sales feedback between April 1, 2024, and July 1, 2024, including 12,696 high-interest customers and 471 contract-signing customers. Using these labeled customers as target nodes, along with their 4-hop neighbors connected through the relations described in Section \ref{sec:data}, we constructed a heterogeneous graph comprising over 6 million nodes and 10 million edges.
Our goal is to predict node classification tasks: 1) High-Interest Customer prediction; 2) Contract-Signing Customer prediction. 

\textbf{DBLP} \cite{HGB} and \textbf{Aminer} \cite{HetSANN}. They are two widely-known academic citation networks in computer science.  
DBLP contains three node types (author, paper, and phase) and two node classification tasks: 1) Author research field prediction;
2) Paper publication venue prediction.
Aminer contains two node types (author and paper) and two node classification tasks: 1) Author research field prediction; 2) Paper research field prediction.

\begin{table}  
\caption{Dataset Information}
\centering

\begin{tabular}{@{}llllll@{}}
\toprule
Dataset & \# Nodes & \begin{tabular}[c]{@{}l@{}}\# Node\\  Types\end{tabular} & \# Edge & \begin{tabular}[c]{@{}l@{}}\# Edge \\ Types\end{tabular} & \# Task \\ \midrule
DBLP & 26,108 & 3 & 223,978 & 6 & 2 \\
Aminer & 28,253 & 2 & 139,831 & 4 & 2 \\
Logistic & 6,142,839 & 4 & 11,592,443 & 5 & 2 \\ \bottomrule
\end{tabular}

\label{tb:dataset_info}
\end{table}

\begin{table*}[ht]
\centering
\caption{Model performance. Each model is trained 5 times with different seeds. The best results are marked in bold. }
\begin{tabular}{@{}lllllllll@{}}
\toprule
\multicolumn{1}{c|}{Dataset} & \multicolumn{4}{c|}{Aminer} & \multicolumn{4}{c}{DBLP} \\ \midrule
\multicolumn{1}{c|}{Tasks} & \multicolumn{2}{c}{task1} & \multicolumn{2}{c|}{task2} & \multicolumn{2}{c}{task1} & \multicolumn{2}{c}{task2} \\
\midrule
\multicolumn{1}{c|}{Metric} & \multicolumn{1}{c}{micro F1} & \multicolumn{1}{c}{macro F1} & \multicolumn{1}{c}{micro F1} & \multicolumn{1}{c|}{macro F1} & \multicolumn{1}{c}{micro F1} & \multicolumn{1}{c}{macro F1} & \multicolumn{1}{c}{micro F1} & \multicolumn{1}{c}{macro F1} \\
\midrule
\multicolumn{1}{c|}{STL-HGT} & $90.03_{\pm 0.50}$ & $90.05_{\pm0.49}$ & $92.51_{\pm0.17}$ & \multicolumn{1}{c|}{$92.57_{\pm0.21}$} & $82.67_{\pm0.76}$ & $81.82_{\pm0.87}$ & $35.39_{\pm0.47}$ & $18.63_{\pm1.07}$ \\
\multicolumn{1}{c|}{STL-HGB} & $91.16_{\pm0.19}$ & $91.64_{\pm0.19}$ & $92.58_{\pm0.22}$ & \multicolumn{1}{l|}{$92.62_{\pm0.24}$} & $85.51_{\pm0.74}$ & $84.67_{\pm0.78}$ &  $37.39_{\pm0.20}$ &  $19.10_{\pm1.60}$ \\
\midrule
\multicolumn{1}{c|}{Shared-HGT} & $93.35_{\pm0.47}$ & $93.78_{\pm0.44}$ & $93.38_{\pm0.24}$ & \multicolumn{1}{c|}{$93.41_{\pm0.23}$} & $84.71_{\pm0.49}$ & $83.59_{\pm0.46}$ & $32.90_{\pm0.51}$ & $12.86_{\pm1.02}$ \\
\multicolumn{1}{c|}{Shared-HGB} &  $93.89_{\pm0.28}$ &  $94.28_{\pm0.25}$ &  $93.41_{\pm0.18}$ & \multicolumn{1}{l|}{ $93.48_{\pm0.16}$} & $91.02_{\pm0.36}$ & $90.34_{\pm0.27}$ & $36.34_{\pm0.98}$ & $14.21_{\pm1.87}$ \\
\multicolumn{1}{c|}{MMOE} & $93.63_{\pm0.66}$ & $93.86_{\pm0.64}$ & $93.20_{\pm0.28}$ & \multicolumn{1}{l|}{$93.24_{\pm0.28}$} & $91.01_{\pm0.23}$ & $90.29_{\pm0.25}$ & $35.18_{\pm0.57}$ & $13.12_{\pm0.75}$ \\
\multicolumn{1}{c|}{PLE} & $93.72_{\pm0.13}$ & $94.13_{\pm0.14}$ & $92.96_{\pm0,33}$ & \multicolumn{1}{c|}{$93.07_{\pm0.35}$} &  $91.22_{\pm0.30}$ & $90.38_{\pm0.33}$ & $35.59_{\pm0.98}$ & $14.23_{\pm1.29}$ \\
\multicolumn{1}{c|}{MultiSFS} & $93.47_{\pm0.30}$ & $94.17_{\pm0.28}$ & $93.19_{\pm0.38}$ &  \multicolumn{1}{c|}{$93.16_{\pm0.35}$} & $91.10_{\pm0.13}$ & $90.47_{\pm0.14}$ & $36.14_{\pm0.69}$ & $14.56_{\pm0.99}$ \\
\midrule
\multicolumn{1}{c|}{StrucHIS} & $\pmb{95.01_{\pm0.41}}$ & $\pmb{95.34_{\pm0.38}}$ & $\pmb{93.52_{\pm0.19}}$ & \multicolumn{1}{c|}{$\pmb{93.54_{\pm0.37}}$} & $\pmb{91.43_{\pm0.28}}$&  $\pmb{90.56_{\pm0.39}}$ & $\pmb{38.25_{\pm0.87}}$ & $\pmb{20.92_{\pm1.14}}$ \\ \bottomrule
\end{tabular}

\label{tab:result_pb}
\end{table*}
\begin{table}[t!]
\centering

\caption{Model performance. Each model is trained 5 times with different seeds. The best results are marked in bold. }
\begin{tabular}{@{}lllllll@{}}
\toprule
\multicolumn{1}{c|}{Dataset} & \multicolumn{4}{c}{Logistics Dataset } \\ 
\midrule
\multicolumn{1}{c|}{Tasks} & \multicolumn{2}{c}{High-Interest} & \multicolumn{2}{c}{Contract-Signing} \\
\midrule
\multicolumn{1}{c|}{Metric} & \multicolumn{1}{c}{AUC} & \multicolumn{1}{c}{AP} & \multicolumn{1}{c}{AUC} & \multicolumn{1}{c}{AP}  \\
\midrule
\multicolumn{1}{c|}{STL-HGT} & $60.23_{\pm0.17}$ & $4.27_{\pm0.04}$ &  $56.11_{\pm0.19}$ & $1.02_{\pm0.03}$ \\
\multicolumn{1}{c|}{STL-HGB} &  $61.56_{\pm0.25}$ &  $4.51_{\pm0.08}$ &  $58.91_{\pm0.21}$ &  $1.42_{\pm0.02}$  \\
\midrule
\multicolumn{1}{c|}{Shared-HGT} & $59.09_{\pm0.40}$ & $4.11_{\pm0.11}$ &  $56.10_{\pm0.34}$ & $1.07_{\pm0.09}$  \\
\multicolumn{1}{c|}{Shared-HGB}  & $60.38_{\pm0.33}$ & $4.31_{\pm0.09}$ & $56.45_{\pm0.28}$ & $1.12_{\pm0.06}$  \\
\multicolumn{1}{c|}{MMOE} & $60.46_{\pm0.74}$ & $4.38_{\pm0.17}$ & $56.98_{\pm0.53}$ & $1.25_{\pm0.12}$ \\
\multicolumn{1}{c|}{PLE} & $60.51_{\pm0.88}$ & $4.42_{\pm0.21}$ & $57.36_{\pm0.87}$ & $1.31_{\pm0.17}$  \\
\multicolumn{1}{c|}{MultiSFS }& $60.74_{\pm0.65}$ & $4.49_{\pm0.19}$ & $57.32_{\pm0.91}$ & $1.29_{\pm0.15}$  \\
\midrule
\multicolumn{1}{c|}{StrucHIS} & $\pmb{65.21_{\pm0.49}}$ & $\pmb{5.67_{\pm0.14}}$  & $\pmb{63.44_{\pm0.43}}$ & $\pmb{2.15_{\pm0.10}}$ \\ 
\bottomrule
\end{tabular}

\label{tab:result_jd}
\end{table}
\subsection{Experimental Settings}

\hspace{3mm}\textbf{Baseline Methods.}\;
We compare our method with the following mainstream and competitive baseline methods, which can be classified into three categories:
\begin{itemize}[leftmargin=*]
    \item \textbf{Single-task Learning (STL) methods:} We choose two representative HGNNs, \textbf{STL-HGT} \cite{HGT} and \textbf{STL-HGB} \cite{HGB}, as our base models. Under this setting, each task is independently trained. 
    \item \textbf{Shared-Backbone MTL methods:}  
    \textbf{MTL-HGT}, \textbf{MTL-HGB} use a shared HGNN for all tasks with separate prediction heads for each task.
    \item \textbf{MoE-based MTL methods:} \textbf{MMoE} \cite{MMoE} uses multiple "expert" networks to extract diverse information and gate networks to dynamically fuse them.
\textbf{PLE}~\cite{PLE} assigns task-specific and shared experts to push them to learn specific information.   
\textbf{MultiSFS}~\cite{MultiSFS} further employs task-specific feature selection.   
    Since these methods are all implemented for the learning of non-graph data, they utilize feed-forward neural networks to implement their experts. \textbf{In our scenario, we implement their experts with HGNNs to ensure an equitable comparison with our method}. Here, we choose HGB, as it achieves better results than HGT.
\end{itemize}

\textbf{Implementations.}\;
For a fair comparison, we implement all methods with the same number of graph layers: 4 for Logistics and 3 for DBLP and Aminer. The hidden layer dimensions are also consistent: 128 for the Logistics, and 64 for DBLP and Aminer.

We divide all datasets into train, validation, and test sets. For the Logistic dataset, we split by time (0-70 days for training, 70-80 days for validation, and 80-90 days for testing). For DBLP and Aminer, we split randomly (6:2:2 ratio). 
Models are trained on the training set, halted when the model's performance on the validation set no longer improves after 40 epochs, and then evaluated on the test set.
As the Logistic dataset is too large, we use graph sampling~\cite{HGT} to sample a sub-graph for training in each epoch. We ensure the ratio of positive samples is 0.5 in after graph sampling. We input the whole graph of DBLP and Aminer into the model for training. 
We use the Adam optimizer~\cite{Adam} for training and perform a grid search for the learning rate and weight decay rate to optimize model performance. 
For the Logistic dataset, we use AUC (area under the ROC curve) and AP (Average Precision). 
For DBLP and Aminer datasets, we follow previous works ~\cite{HGB,HGN-Survey,HetSANN} to evaluate the node classification task with Micro F1 and Macro F1.

\subsection{Logistics Dataset Result Analysis} \label{sec:exp_results}
\begin{itemize}[leftmargin=*]
    \item \textbf{STL methods:}
    As shown in Table \ref{tab:result_jd}, the Contract-Signing Customer prediction task suffers from severe positive label sparsity, leading to low AUC and AP when trained independently.
    \item \textbf{Existing MTL methods vs. STL methods:}
     Ideally, MTL models should get better results on all tasks than STL models. However, both Shared-backbone MTL and MoE-based MTL methods show a significant performance drop on all tasks, highlighting the presence of structural interference across tasks.
    While MoE-based methods slightly outperform Shared-Backbone MTL, they still fail to mitigate structural interference effectively.
    \item \textbf{Our StrucHIS vs. Existing MTL methods:}
    StrucHIS outperforms existing MTL methods on both tasks and surpasses STL baselines.
    It demonstrates StrucHIS's ability to mitigate structural interference and alleviate the positive label sparsity issue by effective cross-task knowledge sharing, leading to superior predictive performance.
\end{itemize}

\subsection{Public Dataset Results Analysis}
To evaluate the robustness and generalization of our method, we test it on two public heterogeneous graph datasets: DBLP and Aminer. 
We find that the varying importance of structural patterns across tasks is not unique to our Logistics dataset but also occurs in public heterogeneous graph datasets such as DBLP and Aminer (see Appendix \ref{app:challenge_pub} for more details). This indicates that structural interference is a common challenge in multi-task heterogeneous graph learning, which can be effectively addressed by our proposed SturHIS framework.
\begin{itemize}[leftmargin=*]
    \item \textbf{Existing MTL methods vs. STL methods:}
    Compared with STL results, existing MTL methods show performance drop on task 2 in DBLP, indicating the presence of structural interference with severe impact on task 2. MTL methods do not degrade the performance of task 1, as the structural inference may not affect all tasks equally.
    Models may prioritize learning the structural pattern beneficial for the easier task, resulting in the other tasks suffer more from the structural inference.
    \item \textbf{Our StrucHIS vs. Existing MTL methods:}
    Compared with baseline MTL methods, our StrucHIS model bridges the performance drop on task 2 in DBLP while maintaining strong performance on task 1. 
\end{itemize}
\textbf{Why don't baseline MTL methods on Aminer experience performance drop on any tasks? Does this imply that MTL on Aminer is unaffected by structural interference?} Baseline MTL methods on AMiner do not show a performance drop on any tasks, as they benefit more from shared information than they suffer from interference. After addressing structural interference by our StrucHIS, we see continued improvements on Aminer.

\subsection{Ablation Study}
\begin{figure}[!t] 
    \centering
    \captionsetup[subfigure]{aboveskip=-0.5pt}
    \begin{subfigure}[b]{0.23\textwidth}
        \centering
        \includegraphics[width=\textwidth]{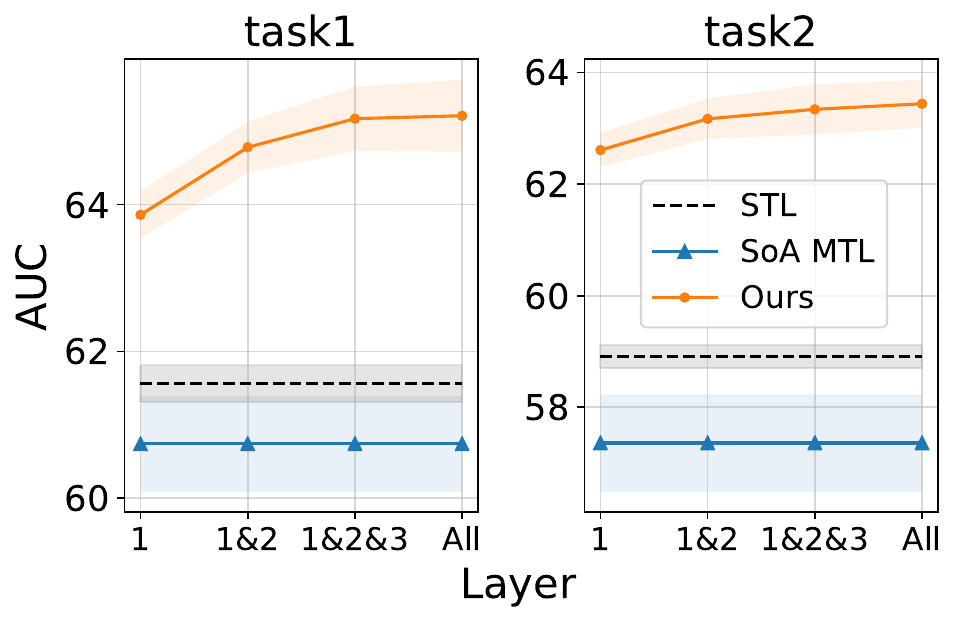}
        \caption{Logistic}
    \end{subfigure}
    \begin{subfigure}[b]{0.23\textwidth}
        \centering
        \includegraphics[width=\textwidth]{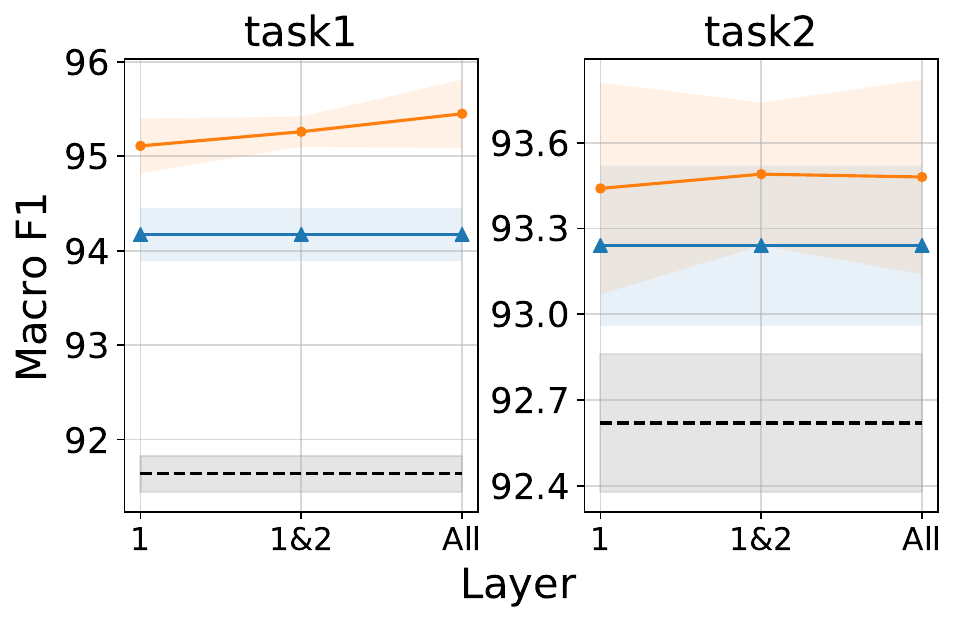}
        \caption{Aminer}
    \end{subfigure}%
    
    \caption{Performance of relation-wise sharing progressively implemented from the first layer onwards. }
\label{fig:app_layer_num}
\end{figure}

In this section, we verify whether it is necessary to selectively share structural information at both relation-level and layer-level. We evaluate the following ablation settings:
\begin{itemize}[leftmargin=*]
    \item \textbf{(w/o R\&L) Without Relation-wise and Layer-wise Sharing.} Tasks use separate HGNN backbones with information sharing only at the final output embedding.
    \item \textbf{(w/o R) Without Relation-wise Sharing.} 
    Relation-specific embeddings are aggregated into the central node, then selective sharing is applied at the aggregated embeddings between tasks at each layer. 
    Table \ref{tb:ablation} shows this setting still suffers performance drops, which indicate that \textbf{layer-wise sharing alone is insufficient and necessitates relation-wise sharing to mitigate structural interference}.
\end{itemize}
We could not build a setting without layer-wise sharing but with relation-wise sharing \textbf{(w/o L)} due to the dependency of relation-wise sharing on layer-wise sharing.
To verify the necessitates to implement information sharing at each layer rather than a single layer, we progressively apply the relation-wise sharing from the first layer onwards. As shown in Figure \ref{fig:app_layer_num}, $1\&2$ denotes that we apply relation-wise sharing in the first two layers,  with no sharing in the third layer.
Results show consistent performance improvements on all datasets, validating the effectiveness of layer-wise sharing. DBLP result is shown in Figure \ref{fig:app_layer_num_DBLP}.

\begin{table}[t]
\centering

\caption{Ablation Study. The metric for Aminer and DBLP is micro F1, while for Logistic is AUC.}
\begin{tabular}{cccccccc}
\toprule
Dataset              & \multicolumn{2}{c}{Logistic} & \multicolumn{2}{c}{DBLP} & \multicolumn{2}{c}{Aminer} \\
Task                 & Task1        & Task2       & Task1       & Task2      & Task1    & Task2    \\
\midrule
w/o R\&L    &    60.43   &     56.82       &      91.11    &      35.76     &    93.64      &    93.01    \\
w/o R &       60.58       &  57.31 &   91.25      &      36.21       &     93.78       &     93.16      \\
\midrule
StrucHIS       &     \textbf{65.21}         &     \textbf{63.44}        &      \textbf{91.43}      &     \textbf{38.25}       &     \textbf{95.01}     &       \textbf{93.52}\\  
\bottomrule
\end{tabular}

\label{tb:ablation}
\end{table}

\subsection{Computation Efficiency Analysis}\label{app:time}
Let $n$, $|E|$, $|\TB_{\VM}|$, and $|\TB_{\EM}|$ denote the number of nodes, edges, node types, and edge types, $L$ denotes the number of HGNN layers, $T$ denotes the number of tasks. Assume all representations have the same dimension $d$.  The time complexity of Shared-Backbone MTL methods is $O(HGNN)$. MoE-based MTL methods have similar time complexity $(T+1) O(HGNN)+T O(gate)$. Our StrucHIS's time complexity is $T O(HGNN)+L \frac{|\TB_{\EM}|}{|\TB_{\VM}|} TO(gate)$.  
The complexity difference between StrucHIS and MoE-based MTL methods mainly lies in the number of HGNNs and gates. 
Given that $T \ll d \ll n \le |E|$, $O(gate)=O(nTd)$ is much smaller than $O(HGNN)=O(L(nd^2+|E|d))$. While StrucHIS utilizes more gates, it also reduces the number of HGNNs. As a result, StrucHIS and MoE-based MTL exhibit comparable computational costs as shown in Table \ref{tb:time}. Detailed calculation process is provided in the Appendix \ref{app:time}.
To reduce redundancy, we used MTL-HGB and PLE as representative methods for Shared-MTL and MoE-MTL, respectively, since methods within each category exhibit similar computational efficiency.

\begin{table} [!t] 
\centering

\caption{Training time per epoch and Inference time on entire test dataset (second)}
\begin{tabular}{@{}cccc@{}}
\toprule
\begin{tabular}[c]{@{}c@{}}Dataset\end{tabular} & \begin{tabular}[c]{@{}c@{}}Logistics \end{tabular} & \begin{tabular}[c]{@{}c@{}}DBLP \end{tabular} & \begin{tabular}[c]{@{}c@{}}Aminer\end{tabular}\\ 
\midrule
Shared-MTL & 78.34 | 61.25 & 0.86 | 1.54  &  0.52 | 0.37 \\
MoE-MTL & 261.79 | 80.53  & 2.83 | 3.62&  2.04 | 0.94 \\
Ours & 267.05 | 84.23  & 2.94 | 3.87  &  2.11 | 1.02 \\ 
\bottomrule
\end{tabular}

\label{tb:time}
\end{table}

\begin{figure}[t]
\centering
\includegraphics[scale=0.4]{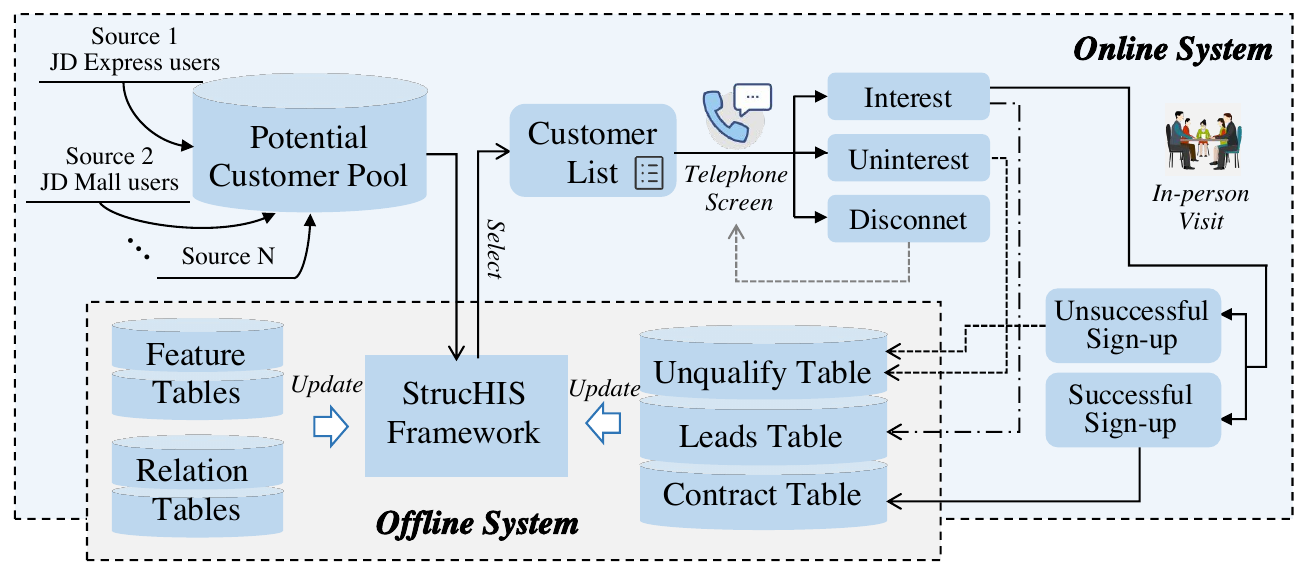}
\caption{Deployment of Our StrucHIS Framework.}
\label{fig:deployment_framework}
\end{figure}

\subsection{\textbf{Real World Deployment}} 
\hspace{3mm}\textbf{System Description:}\; We deployed our model at JD Logistics, one of China’s largest logistics platforms, serving over 500 million customers across more than 360 cities. 
The customer expansion system in JD is shown in Figure \ref{fig:deployment_framework}.
The potential customer pool consists of unconnected customers and is dynamically updated. Our StrucHIS framework estimates contract-signing probabilities of all customers in the pool every two weeks. Customers were then ranked based on these probabilities, and the top $K$ were selected for telephone outreach. 
If a customer shows strong interest in JD Logistics’ services, sales conduct in-person visits to assess key factors such as package volume and shipping unit prices with the customer. Then, a contract was finalized if both parties reached an agreement. During outreach, customers with different level of intention are categorized into different label tables, prompting offline updates to the StrucHIS model every two weeks. 


\textbf{Online A/B Testing:}\;
We conduct an online A/B test to evaluate the effectiveness of StrucHIS. The treatment group utilized our method, while the control group employed two baselines: JD's previous rule-based approach and the single-task HGNN model, HGB.
We collect 431,726 labeled customers based on sales feedback between July 1, 2024, and September 1, 2024, and use these customers and their 4-hop neighbors to construct a logistics heterogeneous graph for model training. For each method, we select and contact 10,000 customers. Since in-person visits to finalize contracts require time, we conducted the final outcome assessment two months after the initial customer outreach.

Table \ref{tb:abtest} shows the number of high-interest customers and contract-signing customers among 10,000 customers selected by each method. Compared to JD's previous method, though STL-HGB achieves slight improvement by leveraging graph learning, its performance is constrained by the positive label sparsity issue. In contrast, our StrucHIS significantly outperforms both methods, which demonstrates the effectiveness of incorporating high-interest tasks to solve the label sparsity through efficient structural information sharing.

Our StrucHIS framework has been fully deployed in JD Logistics and demonstrated substantial commercial value. It successfully identified 18,538 new high-interest customers, of which 845 converted to long-term contract signings, generating over 453K new orders in the past two months.

\begin{table} [t] 
\centering
\caption{Results on online A/B testing.}
\begin{tabular}{@{}cccc@{}}
\toprule
Group &Method & \begin{tabular}[c]{@{}c@{}} \# High-Interest\end{tabular}  & \begin{tabular}[c]{@{}c@{}} \# Contract-Signing\end{tabular} \\ 

\midrule
\multirow{2}{*}{Control}&JD's Method & 254 & 12 \\
& STL-HGB & 272 (+7.1\%)  & 13 (+8.3\%) \\
Treatment& StrucHIS & 336 (+32.3\%) & 17 (+41.7\%) \\ 
\bottomrule
\end{tabular}
\label{tb:abtest}
\end{table}

\section{Related Works}
\subsection{Customer Expansion}
Customer expansion aims to identify new customers who resemble existing “seed” customers, thus broadening campaign reach \cite{Hubble}. 
Existing approaches~\cite{ad1,ad2,ad3} primarily relied on customer attributes (e.g., demographics, basic behaviors) to measure similarity and predict prospective customers.
Recent approaches \cite{Hubble, AD-temporalgraph} recast customer expansion as a single node classification task in a heterogeneous graph. Such a graph-based perspective leverages advanced neural architectures (e.g., attention mechanisms) to capture complex structural patterns around a customer, achieving stronger predictive accuracy and coverage than attribute-only models.
However, these approaches become less accurate when positive samples are extremely sparse.

\subsection{Heterogeneous Graph Representation Learning}
 Heterogeneous graph representation learning aims at learning node representations using node features and graph structures in the heterogeneous graph ~\cite{HGN-Survey, HGN-Survey2}, supporting many downstream tasks like node classification~\cite{hang2024complex,yang2023carpg,Outside-in}, and link prediction~\cite{hang2024paths2pair}.
Compared with homogeneous graph learning, the key challenge in heterogeneous graph representation learning is handling the diversity of node and edge types.
RGCN \cite{RGCN} proposes an edge-specific graph convolution network.
HetSANN \cite{HetSANN} and HGT \cite{HGT} use the attention mechanism to better integrate diverse edge and node information.
Lv et al.~\cite{HGB} establish a comprehensive benchmark for fair comparison of all existing methods and propose the efficient model HGB inspired by their comparisons.

 \subsection{Multi-task learning}
Most MTL methods on heterogeneous graphs use a shared backbone for all tasks to ensure information sharing between tasks, with separate prediction heads for each task ~\cite{Boosting, LZZLD021,SaravanouTML21,LimHNGWT22}.
However, such a shared backbone design leads to performance degradation when tasks require different structural information.  
Although advanced MTL methods on non-graph data employ multiple backbones to implicitly learn shared and specific patterns separately ~\cite{MMoE, PLE, MultiSFS}, they still struggle in heterogeneous graph data. This is due to the inherently complex structure learning process of heterogeneous graph neural networks, which involves multi-layer aggregations of multi-type relations, making it difficult to differentiate the structural dependencies of different tasks.

\section{Conclusion}
In this work, we address a previously under-researched issue:  cross-task structural interference in customer expansion.
To tackle this, we propose StrucHIS, a heterogeneous graph-based MTL framework that explicitly regulates structural knowledge sharing. It breaks down the structure learning phase into multiple stages and introduces sharing mechanisms at each stage, ensuring that task-specific requirements are addressed during each stage.
StrucHIS achieves a 51.41\% AP improvement on private dataset and a 10.52\% macro F1 gain on public datasets. It is further deployed at a leading logistics companies in China and demonstrate a 41.7\% improvement in customer contract-signing rate over existing strategies, generating over 453K new orders within just two months, highlighting its substantial commercial value.

\begin{acks}
Prof. Shuxin Zhong is partially supported by the Guangdong Provincial Key Lab of Integrated Communication, Sensing and Computation for Ubiquitous Internet of Things (No.2023B1212010007), China NSFC Grant (No.62472366), 111 Center (No.D25008), the Project of DEGP (No.2024GCZX003, 2023KCXTD042), Shenzhen Science and Technology Foundation (ZDSYS20190902092853047).
\end{acks}

\bibliographystyle{ACM-Reference-Format}
\bibliography{main}

\appendix

\appendix
\setcounter{secnumdepth}{2}

\section{Appendix}
\subsection{Pseudo-code of StrucHIS} \label{sec:app_method}

\begin{algorithm}[h]
\caption{StrucHIS pipline}\label{alg:cap}
\begin{algorithmic}[1]
\Require A heterogeneous graph $G=\{\VM, \EM, \tau, \psi\}$, StrucHIS model with parameter sets $\theta$, $L$ graph layers, $T$ tasks, and $n$ training epochs.  
\Ensure Prediction results of task 1 to task $T$.
\While{epoch $\leq$ n or not converged}
    \State Feature pre-processing by $h_v = W_{\tau(v)}x_v + b_{\tau(v)}$ and $h_e = W_{\psi(e)}x_e + b_{\psi(e)}$.
    \For{layer $l\in [1,...,L]$}
        \State For each node $v\in \VM$, the structure learning and sharing between tasks is as follows:
        \For{$t \in [1,...,T]$}
        \State 
        $\hat{h_u^{t}}=\mathop{Mes}(h_u^{t,l-1}, \psi(e))$ 
        \For{$\psi_i \in \TB_{\EM}$}
        \State  $h_{\psi_i}^t = \sum\limits_{\forall u, \psi(e_{v,u})=\psi_i}\alpha_u \hat{h_u^t}$ \Comment{Learn relation-specific embedding}
        \EndFor
        \EndFor
        \For{$t \in [1,...,T]$}
        \For{$\psi_i \in \TB_{\EM}$} \Comment{Information Sharing}
        \State 
       $ [w^1,..,w^T] \leftarrow \mathop{Softmax}(\text{FC}^t_{\psi_i}(h_v^{t,l-1}))$
        \State $\overline{h_{\psi_i}^{t}} = \sum_{j=1..T} w^j h_{\psi_i}^{j}$
        \EndFor
        \State $h^{t,l}_v = \sum\limits_{\forall v\in V, \forall \psi_i \in \TB_{\EM}}\left(\beta_{\psi_i} \cdot \overline{h_{\psi_i}^{t}} \right)$
        \Comment{Aggregation across relations}
        \EndFor

    \EndFor
    \For{$t \in [1,...,T]$} \Comment{Prediction on target node $\tilde{v}$}
    \State $\widehat{y^t_{\tilde{v}}} = f_t(h^{t,L}_{\tilde{v}})$
    \EndFor
    \State $L(\theta) = \sum_{t=1}^T\sum_{\tilde{v}} \L_t(y^t_{\tilde{v}}, \widehat{y^t_{\tilde{v}}}).$ \Comment{Model parameter update}

\EndWhile
\end{algorithmic}
\end{algorithm}

\subsection{Datasets} \label{app_dataset}
\paragraph{\textbf{DBLP}}\footnote{https://github.com/THUDM/HGB/tree/master} is an academic citation network.
It contains three nodes types: author, paper, and phase and six edge types. Author is connected to its published papers (a-p, p-a); Paper is connected to the paper it cites (p-p). Paper is connected to phases (term) that can describe this paper (p-ph, ph-p); Author is related to a author (a-a). 
DBLP contains two node classification tasks: 1) Author research field prediction task, where each author is classified to a research field it focuses on, totaling four research fields.
2) Paper publication venue prediction tasks, where each paper is classified to its publication venue, totaling 20 venues.

\paragraph{\textbf{Aminer}}\footnote{https://github.com/didi/hetsann} is also an academic citation network. 
It contains two node types: author and paper and four edge types. Author is connected to its published papers (a-p, p-a); Authors have collaboration relationships with each other (a-a); Paper is connected to the paper it cites (p-p).
Aminer contains two node classification tasks: 1) Author research field prediction task, where each author is classified into multiple research fields it focuses on, totaling 4 research fields. 2) Author research field prediction task, where each paper is classified to a research field it belongs to, totaling 4 research fields.

\subsection{Structural Interference in Public Datasets}\label{app:challenge_pub}
\begin{figure}[htp] 
    \centering
    \captionsetup[subfigure]{aboveskip=-0.5pt}
    \begin{subfigure}[b]{0.23\textwidth}
        \centering
        \includegraphics[width=\textwidth]{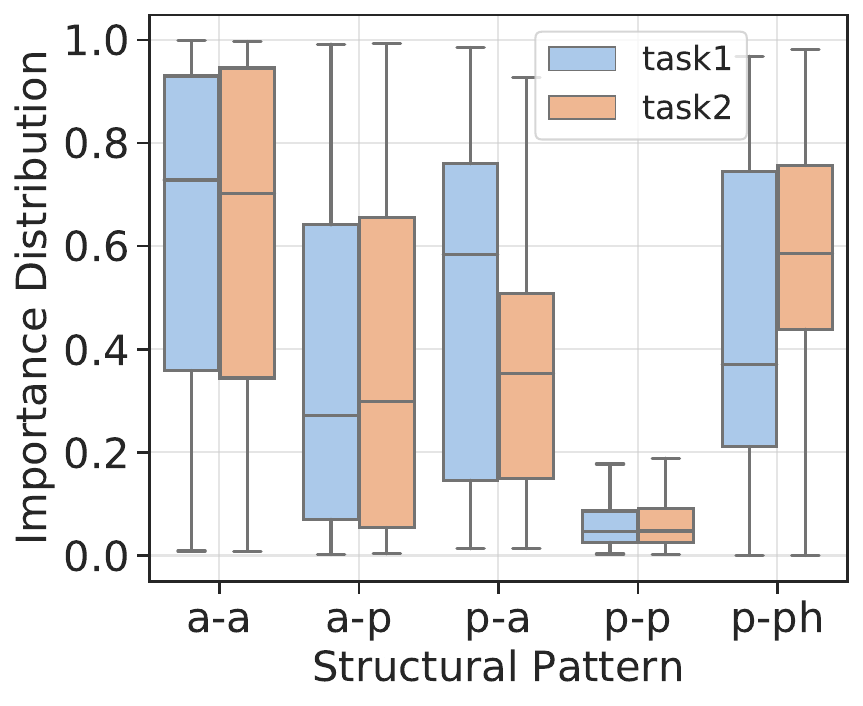}
        \caption{STL Method on DBLP}
    \end{subfigure}
    \hspace{0.9mm} 
    \begin{subfigure}[b]{0.23\textwidth}
        \centering
        \includegraphics[width=\textwidth]{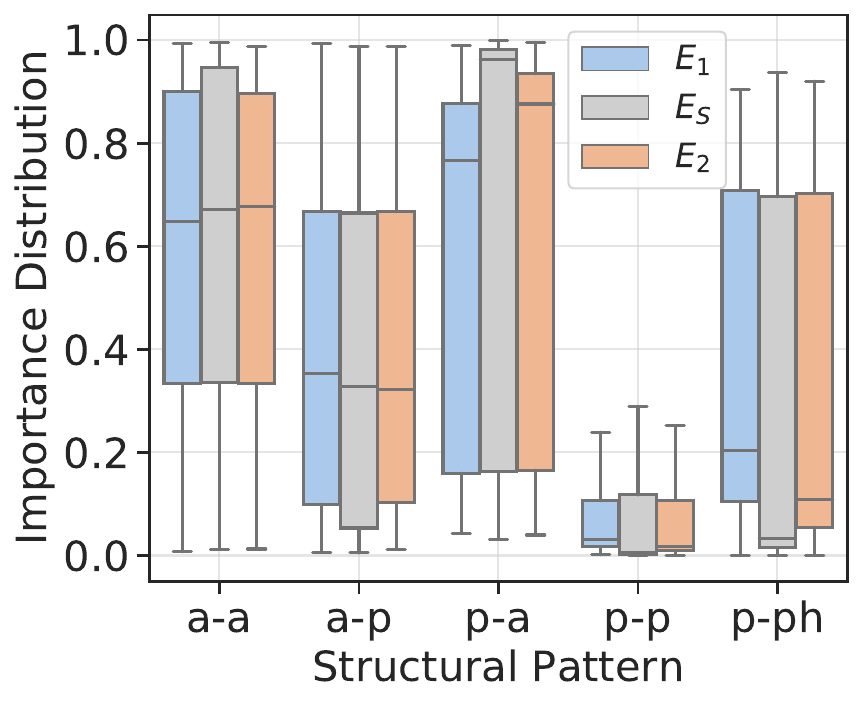}
        \caption{MoE MTL Method on DBLP}
    \end{subfigure}
\caption{(a) The importance distribution of structural patterns
learned by different tasks in STL. (b) The importance distribution of structural patterns learned by different experts ($E_1$, $E_S$, and $E_2$) in MoE-based MTL method PLE. }
\label{fig:DBLP_challenge}
\end{figure}

\begin{figure}[!htp] 
    \centering
    \captionsetup[subfigure]{aboveskip=-0.5pt}
    \begin{subfigure}[b]{0.23\textwidth}
        \centering
        \includegraphics[width=\textwidth]{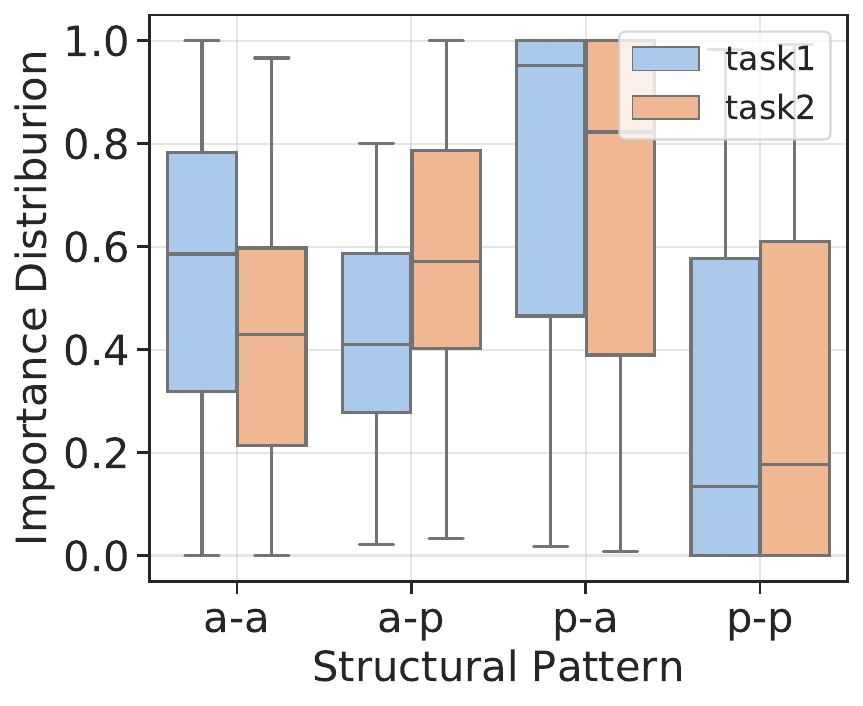}
        \caption{STL Method on Aminer}
    \end{subfigure}
    \hspace{0.9mm} 
    \begin{subfigure}[b]{0.23\textwidth}
        \centering
        \includegraphics[width=\textwidth]{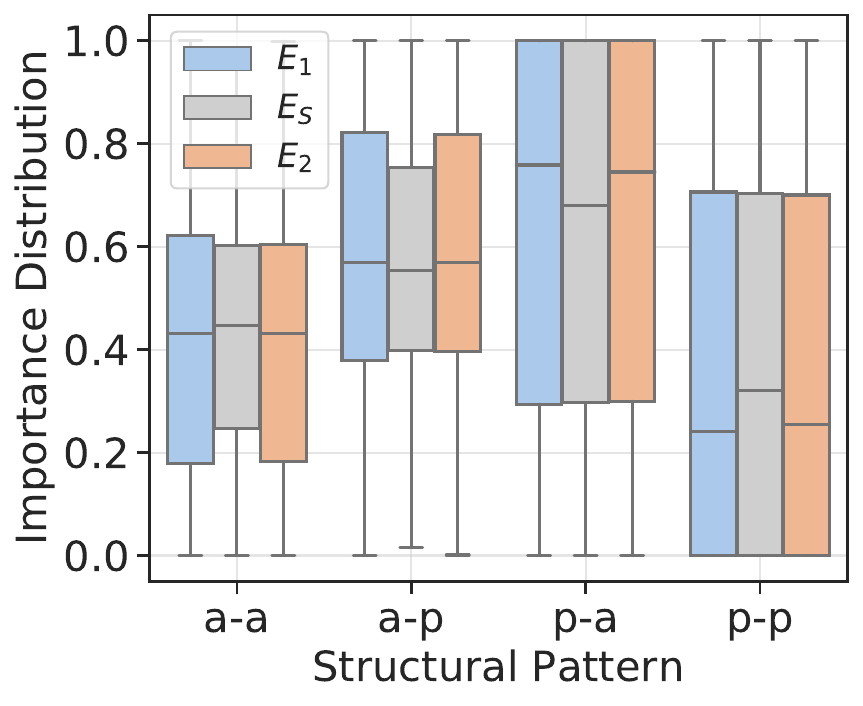}
        \caption{MoE MTL Method on Aminer}
    \end{subfigure}
\caption{Same as Figure~\ref{fig:DBLP_challenge}, but on the Aminer dataset . }
\label{fig:Aminer_challenge}
\end{figure}

\paragraph{\textbf{The varying impact of a specific structural pattern on different tasks.}}
We claim that some structural patterns benefit all tasks, while others are task-specific, leading to performance degradation in MTL.
For DBLP (Figure \ref{fig:DBLP_challenge}(a)), though structural pattern '\textit{a-a}' (\textit{Author-(related)-Author}) is important for both tasks, structural pattern '\textit{p-ph}' (\textit{paper-(relate)-phase}) is significantly more important for Task 1 than for Task 2, 
Such difference across tasks also appears in  Aminer (Figure \ref{fig:Aminer_challenge}(a)), which verifies our claim.

\paragraph{\textbf{Challenges in existing MTL frameworks.}}
SoA MTL methods~\cite{MMoE,PLE, MultiSFS} leverage multiple experts to learn task-specific and shared structural patterns separately. However, our experiment results reveal that they still struggle to discriminate these patterns.
Figure \ref{fig:DBLP_challenge}(b) and \ref{fig:Aminer_challenge}(b) show the structural pattern importance learned by different experts ($E_1$, $E_S$, and $E_2$) in the PLE model~\cite{PLE}. Ideally, $E_1$ and $E_2$ should focus on learning task-specific structural patterns, while $E_S$ should capture commonalities across tasks. However, in the experiments, all experts tend to learn similar and ‘average’ importance for each structural pattern, which demonstrates that PLE struggles to selectively extract task-specific structural information.

\subsection{Computation Efficiency Analysis}\label{app:time}
Let $n$, $|E|$, $|\TB_{\VM}|$, and $|\TB_{\EM}|$ denote the number of nodes, edges, node types, and edge types, $L$ denotes the number of HGNN layers, $T$ denotes the number of tasks. Assume all representations have the same dimension $d$. 
Our time complexity analysis follows a similar fashion as \cite{GAT}:
\begin{itemize} [leftmargin=*]
    \item \textbf{Shared-Backbone MTL}: In one HGNN layer, Message operation needs $O(nd^2)$ and Attention operation needs $O(|E|d)$. Thus, it totally needs $O(HGNN) = O(L(nd^2+|E|d))$.
    \item \textbf{MoE-based MTL}: They use $T+1$ experts and $T$ task-specific gates.
    (1) An expert is a HGNN model, $O(expert)=O(HGNN)$;  
    (2) A gate contains a fully-connected layer to generate $T$ weights for each node, thus $O(gate)=O(nTd)$.    
    Thus, the total complexity is $(T+1)O(HGNN)+TO(gate)$.
    \item \textbf{StrucHIS (Ours)}: We use $T$ HGNN backbones, and $T\cdot L$ gates. A gate contains a fully-connected layer to generate $T\cdot M$ weights for each node. Here, $M$
 denotes the number of edge type surrounding a node, which varies across node types. To simplify, we use the mean value $\frac{|\TB_{\EM}|}{|\TB_{\VM}|}$ to represent $M$. Thus, $O(gate_{StrucHIS})=O(nT\frac{|\TB_{\EM}|}{|\TB_{\VM}|}d)=\frac{|\TB_{\EM}|}{|\TB_{\VM}|}O(gate)$.   
    The total complexity is $TO(HGNN)+L\frac{|\TB_{\EM}|}{|\TB_{\VM}|}TO(gate)$.  
\end{itemize}

The complexity difference between StrucHIS and MoE-based MTL mainly lies in the number of HGNNs and gates. 
Given that $T \ll d \ll n \le |E|$, $O(gate)=O(nTd)$ is much smaller than $O(HGNN)=O(L(nd^2+|E|d))$. While StrucHIS utilizes more gates, it also reduces the number of HGNNs. As a result, StrucHIS and MoE-based MTL exhibit comparable computational costs as shown in Table \ref{tb:time}.

\begin{figure}[htp]
\centering
\includegraphics[scale=.3]{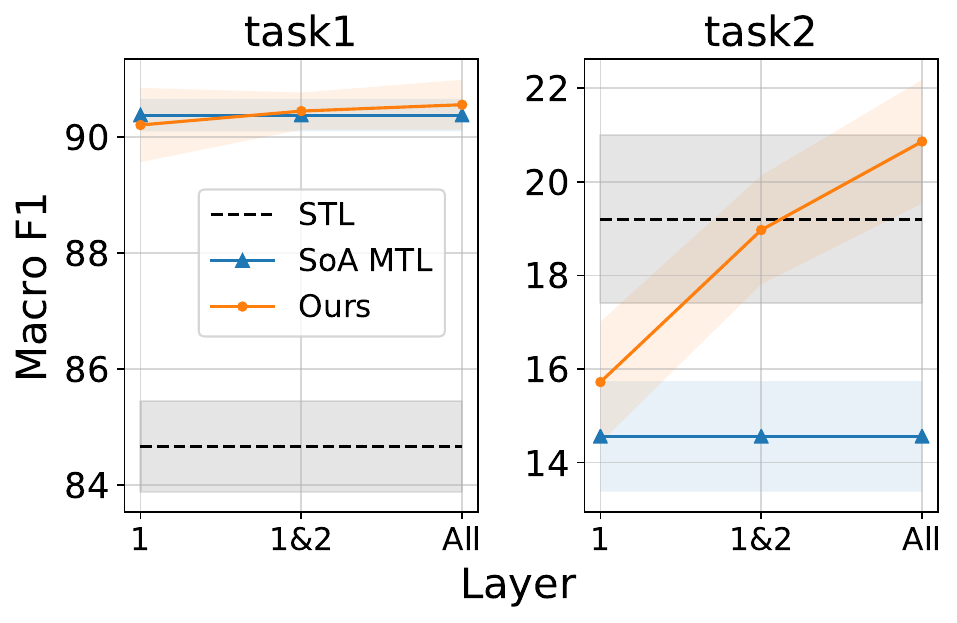}
\caption{Performance of relation-wise sharing progressively implemented from the first layer onwards on DBLP dataset.} 

\label{fig:app_layer_num_DBLP}
\end{figure}

\end{document}